\documentclass[10pt,twocolumn,letterpaper]{article}

\usepackage{cvpr}              

\usepackage{graphicx}
\usepackage{amsmath}
\usepackage{amssymb}
\usepackage{booktabs}
\usepackage[table,xcdraw]{xcolor}
\usepackage{multirow}
\usepackage[misc]{ifsym}

\usepackage[pagebackref,breaklinks,colorlinks]{hyperref}

\usepackage[capitalize]{cleveref}
\crefname{section}{Sec.}{Secs.}
\Crefname{section}{Section}{Sections}
\Crefname{table}{Table}{Tables}
\crefname{table}{Tab.}{Tabs.}

\def\cvprPaperID{1450} 
\def\confName{CVPR}
\def\confYear{2023}

\begin{document}

\title{DeFeeNet: Consecutive 3D Human Motion Prediction with Deviation Feedback}

\author{Xiaoning Sun$^1$, Huaijiang Sun$^1$\textsuperscript{\Letter}, Bin Li$^2$, Dong Wei$^1$, Weiqing Li$^1$, Jianfeng Lu$^1$\\
{$^1$}Nanjing University of Science and Technology, China\\
{$^2$}Tianjin AiForward Science and Technology Co., Ltd., China\\
{\tt\small \{sunxiaoning,sunhuaijiang\}@njust.edu.cn, libin@aiforward.com}
}

\twocolumn[{%
\renewcommand\twocolumn[1][]{#1}%
\maketitle
\begin{center}
    \centering
    \includegraphics[width=.98\textwidth]{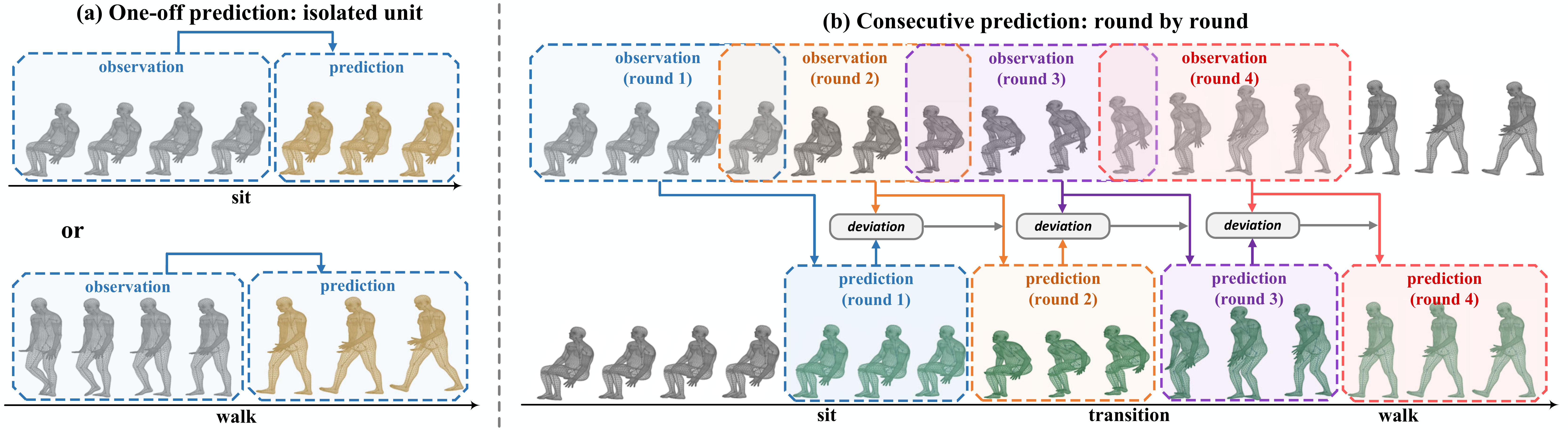}
    \captionof{figure}{(a) Current human motion prediction task: a one-off ``observe then predict'' process within the isolated unit. (b) Consecutive human motion prediction with deviation feedback: covering multiple ``observe then predict'' rounds, where the prediction deviation (i.e., mistake) in previous round could constrain the model to make better prediction in the following round.}
    \label{heading}
\end{center}
}]

\begin{abstract}
  Let us rethink the real-world scenarios that require human motion prediction techniques, such as human-robot collaboration. Current works simplify the task of predicting human motions into a one-off process of forecasting a short future sequence (usually no longer than 1 second) based on a historical observed one. However, such simplification may fail to meet practical needs due to the neglect of the fact that motion prediction in real applications is not an isolated ``observe then predict'' unit, but a consecutive process composed of many rounds of such unit, semi-overlapped along the entire sequence. As time goes on, the predicted part of previous round has its corresponding ground truth observable in the new round, but their deviation in-between is neither exploited nor able to be captured by existing isolated learning fashion.
  In this paper, we propose DeFeeNet, a simple yet effective network that can be added on existing one-off prediction models to realize deviation perception and feedback when applied to consecutive motion prediction task. At each prediction round, the deviation generated by previous unit is first encoded by our DeFeeNet, and then incorporated into the existing predictor to enable a deviation-aware prediction manner, which, for the first time, allows for information transmit across adjacent prediction units. We design two versions of DeFeeNet as MLP-based and GRU-based, respectively. On Human3.6M and more complicated BABEL, experimental results indicate that our proposed network improves consecutive human motion prediction performance regardless of the basic model.
  
\end{abstract}

\section{Introduction}
\label{sec:intro}

In the age of intelligence, humans have to share a common space with robots, machines or autonomous systems for a sustained period of time, such as in human-motion collaboration \cite{chico}, motion tracking \cite{mot-tracking} and autonomous driving \cite{auto-driving} scenarios. To satisfy human needs while keeping their safety, deterministic human motion prediction, which is aimed at forecasting future human pose sequence given historical observations, has become a research hotspot and already formed a relatively complete implement procedure. As human poses may become less predictable over time, current works abstract the practical needs into a simplified task of learning to ``observe a few frames and then predict the following ones'', with the prediction length mostly set to $ \leq $ 1 second \cite{Jain-rnn,Corona-rnn,Fragkiadaki-lstm,Martinez-lstm,seq2seq-cnn,Limaosen-gnn,Mao-2019,Dang-2021,stsgcn,progressively,gating,motmix}.

Essentially, such simplified prediction task can be regarded as a one-off acquired target limited within the short, isolated ``observe then predict'' unit. Nevertheless, this unit is actually not applicable in reality that requires consecutive observation and prediction on humans during the long period of human-robot/machine coexistence.
Though, intuitively, sliding such unit round by round along the time may roughly satisfy the need for consecutive prediction, one neglected fact is that each round of prediction unit is arranged in a \emph{semi-overlapped} structure (see Figure \ref{heading}). As time goes on, 
what was predicted \emph{before} has its corresponding ground truth observable \emph{now},
but their deviation in-between (i.e., mistake) remains unexplored.

Our key insight lies in that robots/machines should be able to detect and learn from the mistakes they have made. 
Due to the inherent continuity and consistency of human motion data, such information would be very powerful to improve future prediction accuracy. The semi-overlapped, multi-round unit structure offers a natural way to transmit deviation feedback across adjacent prediction units, which, however, are unable to be realized by current one-off unit prediction strategy. 


Based on this situation, in this paper, we 
propose DeFeeNet, a simple yet effective network which can be added on existing one-off human motion prediction models to implement \textbf{De}viation perception and \textbf{Fee}dback when applied to consecutive prediction.
By mining the element ``deviation'' that neglected in previous works, we introduce inductive bias, where DeFeeNet can learn to derive certain ``patterns'' from past  deviations and then constrain the model to make better predictions.
To be specific, our DeFeeNet is constructed in two versions: MLP-based version containing one temporal-mixing MLP and one spatial-mixing MLP; GRU-based version containing one-layer GRU with fully-connected layers only for dimension alignment. At each prediction round, DeFeeNet serves as an additional branch inserted into the existing predictor, which encodes the deviation of the previous unit into latent representation and then transmits it into the main pipeline.
In summary, our contribution are as follows:

\begin{itemize}
  \item We mine the element ``deviation'' that neglected in existing human motion prediction works, extending current within-unit research horizon to cross-unit, which, for the first time, allows for information transmit across adjacent units. 
  \item We propose DeFeeNet, a simple yet effective network that can be added on existing motion prediction models to enable consecutive prediction (i) with flexible round number and (ii) in a deviation-aware manner. It can learn to derive certain patterns from past mistakes and constrain the model to make better predictions.
  \item Our DeFeeNet is agnostic to its basic models, and capable of yielding stably improved performance on Human3.6M \cite{dataset-h36}, as well as a more recent and challenging dataset BABEL \cite{dataset-babel} that contains samples with different actions and their transitions.
\end{itemize}

\section{Related Work}
\label{sec:related}

\subsection{Human Motion Prediction}

\noindent\textbf{Model Design Paradigm.} Two mainstreams of model design in human motion prediction are based on sequential networks \cite{Jain-rnn,Corona-rnn,Fragkiadaki-lstm,Martinez-lstm,quaternion-gru} and feed-forward networks \cite{seq2seq-cnn,Limaosen-gnn,Mao-2019,Dang-2021,stsgcn,gating,chico,transformer-3dv,transformer-Cai}. For the former one, RNNs \cite{Jain-rnn,Corona-rnn}, LSTMs \cite{Fragkiadaki-lstm,Martinez-lstm} and GRUs \cite{Martinez-lstm,quaternion-gru,motron} are employed to extract temporal features of motion sequences, while for the latter, CNNs \cite{seq2seq-cnn}, GNNs \cite{Limaosen-gnn}, GCNs \cite{Mao-2019,Dang-2021,stsgcn,gating,chico,sun} and transformers \cite{transformer-3dv,transformer-Cai} learn spatial dependencies or spatio-temporal information to better depict human skeletal structure and joint spatio-temporal connections.
All these current designs simplify human motion prediction as a short, isolated ``observe then predict'' \emph{unit}, with predicted length mostly $ \leq $ 1 second, and with no consideration of its context, which, therefore, could not satisfy practical needs like human-robot collaboration scenario that requires consecutive observation and prediction on human behaviors.

\noindent\textbf{Sample Selection Protocol.} Currently in this field, samples are drawn in a uniform way of cutting small pieces from long motion sequences with known action categories. Each sample piece (i.e., the \emph{unit}) is composed of the first several frames as observation and the others as prediction target, where poses in one piece all belong to the same action category. Such short-length yet single-action sampling convention, therefore, determines that current trained predictors are unable to handle real-world prediction task that covers multiple categories of action and action-changing periods.

Motivated by the above, we reformulate human motion prediction task from the consecutive perspective, and argue that previous prediction deviation should be detected and learned by robots/machines to help the following prediction. We accordingly present a resampling method by drawing multiple yet flexible rounds of \emph{semi-overlapped} unit as \emph{one} sample with a relatively long horizon (Section \ref{3.4} for details), which enables (i) our deviation-aware consecutive prediction as well as (ii) the first attempt to consider action-changing in deterministic human motion prediction. Note that in this paper we only refer to \emph{deterministic} motion prediction that aims to predict only \emph{one} future sequence. Stochastic prediction (predicting multiple possible sequences) \cite{Ali-stocha,dlow,Mao-2021,Mao-2022,wei} or motion synthesis (generating sequences without observations) \cite{syn-act2mot,syn-actor} are not within the scope of our discussion.

\subsection{Deviation-Aware Prediction}
\label{2.2}

Other seemingly similar ideas on deviation-aware prediction include Local-Behavior-Aware (LBA) framework in trajectory prediction \cite{LBA} and Residual Correction framework in node regression \cite{residual-regression}, node classification \cite{residual-classification} and traffic prediction \cite{residual-dest,residual-kaist}. For the former, LBA \cite{LBA} refers to involving a collection of all historical trajectories at the object's current location to help narrow down the search space of future trajectories. Though the ``historical trajectories of other objects'' notion somewhat resembles our ``previous prediction round'', it is entirely unrelated to the use of deviation. For the latter, residual is defined as the difference between GT and the prediction just like our deviation, but all these works first generate an ``original'' prediction, which is then \emph{directly} added with the estimated residual, so that the original prediction could be corrected to some extent (note that the \emph{estimated} residual is calculated based on residuals of training vertices \cite{residual-regression,residual-classification} or based on historical residuals \cite{residual-dest,residual-kaist}). In other words, their residual correction serves as a post-processing step, and the residual expression is completely independent of the original pipeline. In contrast, we adopt the ``encoder-separated while decoder-unified'' strategy, which allows for separate coding on previous deviation while joint decoding on both information of current observation and previous deviation. Therefore, our prediction is naturally improved rather than corrected by post-processing, and that is why we regard it as deviation-aware prediction but not prediction correction. As human motion data is more \emph{granular} yet delicate that cannot be corrected by simply adding a residual/deviation vector, our joint decoding strategy could leverage the decoding power of the original predictor, enabling deviation-aware prediction only with a lightweight network branch inserted.

\section{Proposed Method}
\label{method}

In most cases, 3D human structure is represented by pure pose parameters based on human joint coordinates or SMPL model \cite{smpl} which depicts 3D human mesh with both pose and shape. According to task demand, we follow \cite{syn-actor,Mao-2022} to discard the shape term and only predict the pose term. Currently, human motion prediction \cite{Mao-2019,stsgcn,motmix} aims to predict a future sequence $ \hat{\textbf{Y}} = [\hat{\textbf{y}}_1, \hat{\textbf{y}}_2, \cdots, \hat{\textbf{y}}_T ] $ with $ T $ frames based on the observed $ N $-frame $ \textbf{X} = [\textbf{x}_1, \textbf{x}_2, \cdots, \textbf{x}_N ] $ sequence, where each frame (i.e., pose) is represented by $ \hat{\textbf{y}}_i \in \mathbb{R} ^K $ and $ \textbf{x}_i \in \mathbb{R} ^K $, respectively. We define such process of predicting $T$ frames with the $N$-frame observation as \emph{unit}. Research within this isolated unit, however, totally neglects the importance of ``prediction mistake'' element to consecutive motion prediction, and is unable to realize deviation perception and feedback to improve prediction accuracy.

\subsection{Reformulation from Consecutive Perspective}
\label{3.1}

We introduce our reformulated human motion prediction task. Consecutive prediction involves multiple rounds of unit with $ N+T $ frames semi-overlapped along the entire sequence. Suppose a sequence $ \textbf{S} = [\textbf{s}_1, \textbf{s}_2, \cdots, \textbf{s}_L] $ with length $ L $ long enough, we define the first round as $ \textbf{R}_1=\textbf{s}_{1:N+T} $, and every following round is $ T $ frames further, i.e., the $ r $-th round $ \textbf{R}_r=\textbf{s}_{1+(r-1)T:N+T+(r-1)T} $, where $ r\geqslant1 $, and in this paper we set $ N \geqslant T $. Therefore, if we define all the motion rounds $ \textbf{R} = [\textbf{R}_1, \textbf{R}_2, \cdots, \textbf{R}_r, \cdots] $ and rewrite $ \textbf{R}_r = [\textbf{X}_r, \textbf{Y}_r] = [\textbf{x}_{r,1}, \cdots, \textbf{x}_{r,N}, \textbf{y}_{r,1}, \cdots, \textbf{y}_{r,T}] $, it can be deduced that $ \textbf{x}_{r,N-T:N} = \textbf{y}_{r-1,1:T} $, and the deviation generated in round $ \textbf{R}_{r-1} $ is expressed as $ \textbf{D}_{r-1} = f_d(\textbf{x}_{r,N-T:N}, \hat{\textbf{y}}_{r-1,1:T}) $, to be used to improve the prediction accuracy of round $ \textbf{R}_r $. When $r=1$, $\textbf{D}_0$ is empty as no deviation can be used. Basically, we regard the task of consecutive motion prediction as implement $ \textbf{R} $ round by round.

\begin{figure*}[t]
  \centering
   \includegraphics[width=0.98\linewidth]{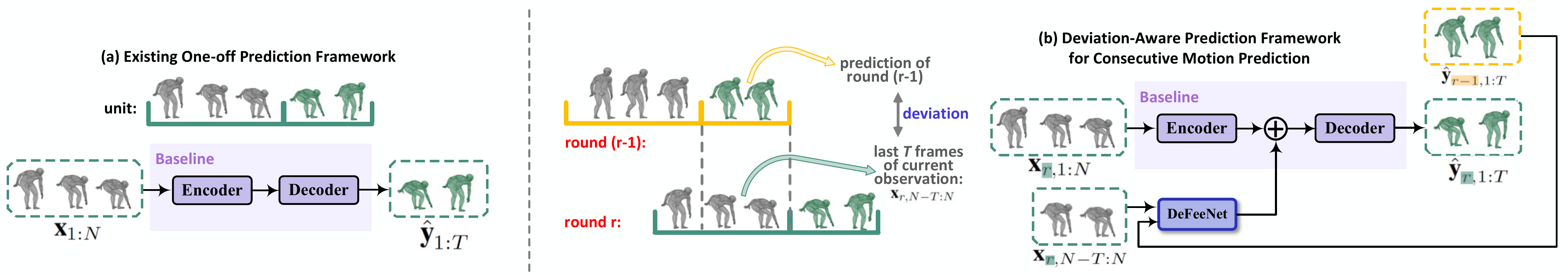}

   \caption{(a) Existing baselines that simplify human motion prediction as a one-off process of predicting a $ T $-pose sequence based on $ N $-pose observation (i.e., the isolated prediction unit). (b) DeFeeNet allows for human motion prediction \emph{round by round} with a deviation-aware manner. Round $ r $ is marked in mint green while $ r-1 $ in orange. During the prediction of round $ r $, our DeFeeNet detects the deviation between current observation and previous round of prediction, to realize deviation feedback across adjacent units.}
   \label{fig:backbone}
\end{figure*}

\begin{figure}[t]
  \centering
   \includegraphics[width=0.84\linewidth]{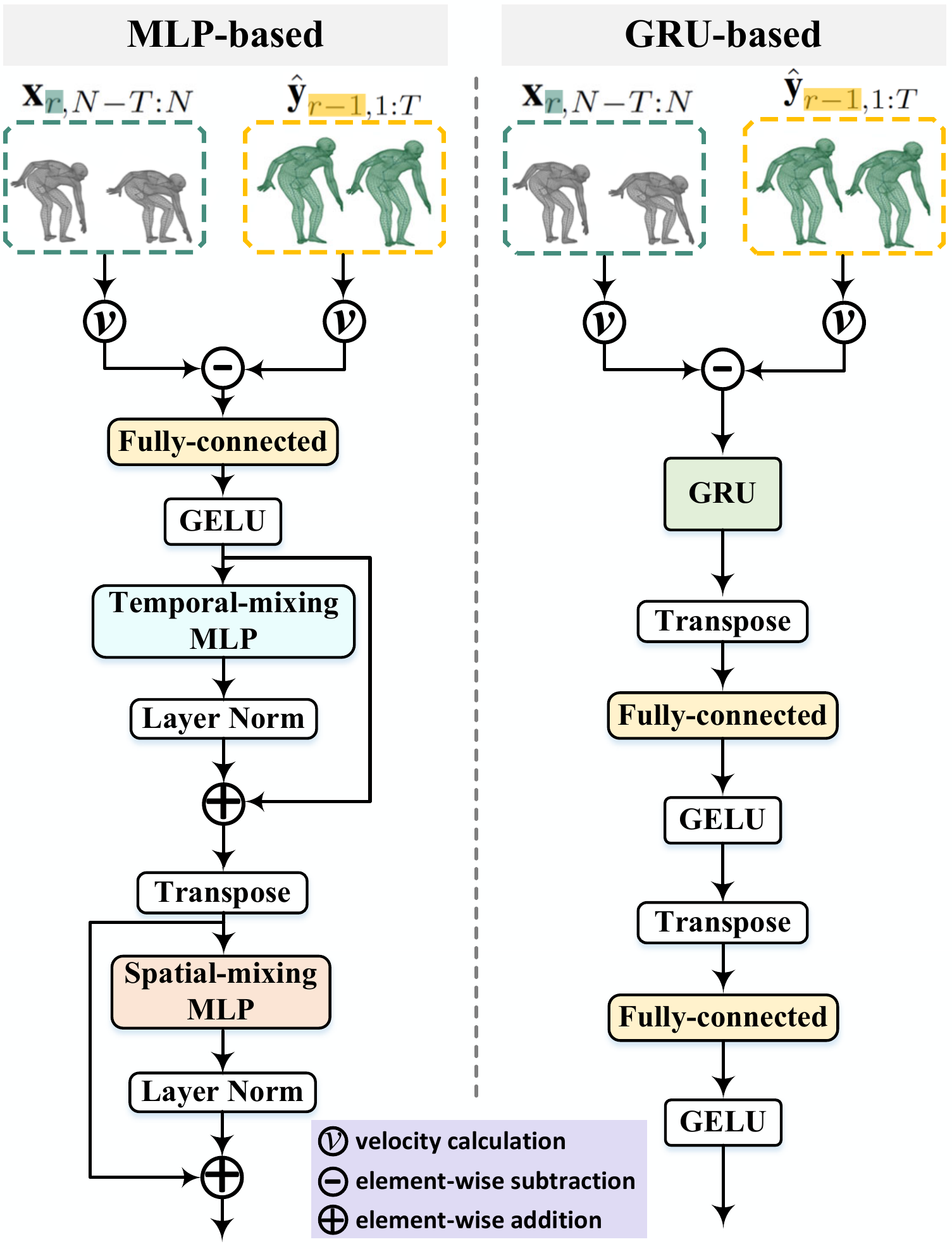}

   \caption{Architectures of MLP-based and GRU-based DeFeeNet.}
   \label{fig:backbone-detail}
\end{figure}

\subsection{Deviation-Aware Prediction}

To express deviation $ \textbf{D}_{r-1} = f_d(\textbf{x}_{r,N-T:N}, \hat{\textbf{y}}_{r-1,1:T}) $ mentioned in Section \ref{3.1} in detail, we draw inspiration from \cite{motmix,motmlp} that design their prediction loss function based on joint displacement of two adjacent poses (i.e., velocity) rather than joint position. Instead of directly subtracting between $ \textbf{x}_{r,N-T:N} $ and $ \hat{\textbf{y}}_{r-1,1:T} $, we first calculate their velocities as $ v(\textbf{x}_{r,N-T:N}) $, $ v(\hat{\textbf{y}}_{r-1,1:T}) \in \mathbb{R} ^{K \times (T-1)} $ and then obtain:
\begin{equation}
  \begin{aligned}
  \textbf{D}_{r-1} &= f_d(\textbf{x}_{r,N-T:N}, \hat{\textbf{y}}_{r-1,1:T})\\
  &= v(\textbf{x}_{r,N-T:N}) - v(\hat{\textbf{y}}_{r-1,1:T}) \in \mathbb{R} ^{K \times (T-1)}.
  \label{D_r-1}
  \end{aligned}
\end{equation}

In this way, compared to existing works that focus on learning a prediction function $ \textbf{Y} = \varphi_o(\textbf{X}) $ between the observation $ \textbf{X} $ and the target $ \textbf{Y} $ in the isolated unit, our deviation-aware consecutive prediction aims to find a function for every prediction round with deviation feedback incorporated:
\begin{equation}
  \textbf{Y}_r = \varphi(\textbf{X}_r, \textbf{D}_{r-1}),
  \label{formula}
\end{equation}
where $ \varphi:\mathbb{R} ^{K \times N} \times \mathbb{R} ^{K \times (T-1)}\mapsto \mathbb{R} ^{K \times T} $. Note that $ \textbf{D}_0 = \textbf{0} $ when $ r=1 $.

In Figure \ref{fig:backbone}, we provide an overview of how DeFeeNet is embedded into the existing prediction framework and helps implement deviation-aware motion prediction round by round during a relatively long period. At round $ r $, we feed the current round of observation $ \textbf{X}_r = \textbf{x}_{r,1:N} $ into the main pipeline (i.e., the existing predictor). Meanwhile, our DeFeeNet could detect the prediction deviation that \emph{just happened} as the prediction of round $ r-1 $ (i.e., $ \hat{\textbf{Y}}_{r-1} = \hat{\textbf{y}}_{{r-1},1:T} $) has its ground truth observable now. Therefore, with this deviation $ \textbf{D}_{r-1} $ coded and incorporated, the current prediction $ \hat{\textbf{Y}}_r = \hat{\textbf{y}}_{r,1:T} $ could be naturally improved.

\subsection{Network Architecture}

We build two versions of DeFeeNet based on MLP and GRU (see Figure \ref{fig:backbone-detail}), both of which are simple yet effective, as we do not encourage DeFeeNet to increase the computation burden too much on original prediction models.

Inspired by the success of MLP-Mixer structure \cite{mlpmixer} in computer vision and its variants in downstream human motion modeling tasks \cite{motmix,motmlp,mlp-generation}, we propose an MLP-based version of our DeFeeNet, but is more lightweight yet effective enough. Compared to using multiple MLP-mixer blocks in \cite{mlpmixer,motmix,motmlp,mlp-generation}, our design discards block stacking and only contains one temporal-mixing MLP, one spatial-mixing MLP, and skip-connections (see Figure \ref{fig:backbone-detail} left). Both MLP-mixer module are composed of two fully-connected layers, GELU activation function \cite{gelu}, and layer normalization operation \cite{layernorm}.

For the GRU-based version, as is shown in Figure \ref{fig:backbone-detail} right, we employ a one-layer GRU attached with one fully-connected layer in temporal dimension and one in spatial dimension for dimension alignment, to be inserted into the original pipeline for following calculation.

To be specific, the original predictor is first fed with the observed segment $ \textbf{X}_1 = \textbf{x}_{1,1:N} $ in $ \textbf{R}_1 $, and obtain its first round of prediction $ \hat{\textbf{Y}}_1 = \hat{\textbf{y}}_{1,1:T} $ at timestep $ T $, with DeFeeNet nonactivated and no deviation information involved. However, during the following rounds where $ \textbf{X}_r $ becomes observable, the deviation between $ v(\hat{\textbf{y}}_{r-1,1:T}) $ and $ v(\textbf{x}_{r,N-T:N}) $ appears, and is fed into our DeFeeNet to capture its temporal and spatial information to be added into the main pipeline, enabling deviation feedback across adjacent prediction units.

\begin{table*}[ht]
  \centering
  \renewcommand{\arraystretch}{1.1}
  \setlength\tabcolsep{3.2pt}
  \scalebox{0.7}{
    \begin{tabular}{c|cccc|cccc|cccc|cccc|cccc}
      \hline
       & \multicolumn{4}{c|}{walking} & \multicolumn{4}{c|}{eating} & \multicolumn{4}{c|}{smoking} & \multicolumn{4}{c|}{discussion} & \multicolumn{4}{c}{directions} \\
      frame num. & 2 & 4 & 8 & 10 & 2 & 4 & 8 & 10 & 2 & 4 & 8 & 10 & 2 & 4 & 8 & 10 & 2 & 4 & 8 & 10 \\ \hline
      LTD-GCN \cite{Mao-2019} & 12.3 & 23.0 & 39.8 & 46.1 & 8.4 & 16.9 & 33.2 & 40.7 & 8.0 & 16.2 & 31.9 & 38.9 & 12.5 & 27.4 & 58.5 & 71.7 & 9.0 & 19.9 & 43.4 & 53.7 \\ \hline
      LTD-DeFee(MLP)-r1 & 11.7 & 21.2 & 36.4 & 43.5 & 8.1 & 16.2 & 31.9 & 39.8 & 7.7 & 15.3 & 29.9 & 36.8 & 12.7 & 26.8 & 56.4 & 69.4 & 9.1 & 19.3 & 41.7 & 52.0 \\
      \rowcolor[HTML]{E8E9FF} 
      LTD-DeFee(MLP)-r2 & \textbf{10.2} & \textbf{19.9} & \textbf{35.5} & \textbf{42.5} & \textbf{6.7} & \textbf{14.8} & \textbf{30.7} & \textbf{38.3} & \textbf{6.5} & \textbf{14.0} & \textbf{28.8} & \textbf{35.6} & \textbf{10.0} & \textbf{23.8} & \textbf{54.5} & \textbf{68.1} & \textbf{6.9} & \textbf{16.7} & \textbf{39.6} & \textbf{50.2} \\
      LTD-DeFee(GRU)-r1 & 11.8 & 22.2 & 36.8 & 44.0 & 8.2 & 16.9 & 32.2 & 40.2 & 7.8 & 15.8 & 30.2 & 37.5 & 12.5 & 26.7 & 57.1 & 70.4 & 8.8 & 19.5 & 42.4 & 52.8 \\
      \rowcolor[HTML]{E8E9FF} 
      LTD-DeFee(GRU)-r2 & \textbf{10.4} & \textbf{20.0} & \textbf{34.7} & \textbf{42.2} & \textbf{7.0} & \textbf{15.2} & \textbf{31.4} & \textbf{38.4} & \textbf{6.8} & \textbf{14.5} & \textbf{29.0} & \textbf{35.8} & \textbf{11.1} & \textbf{25.4} & \textbf{55.8} & \textbf{68.2} & \textbf{7.0} & \textbf{17.0} & \textbf{40.0} & \textbf{50.9} \\ \hline \hline
       & \multicolumn{4}{c|}{greeting} & \multicolumn{4}{c|}{phoning} & \multicolumn{4}{c|}{posing} & \multicolumn{4}{c|}{purchases} & \multicolumn{4}{c}{sitting} \\
      frame num. & 2 & 4 & 8 & 10 & 2 & 4 & 8 & 10 & 2 & 4 & 8 & 10 & 2 & 4 & 8 & 10 & 2 & 4 & 8 & 10 \\ \hline
      STS-GCN \cite{stsgcn} & 18.7 & 34.9 & 71.6 & 86.4 & 13.7 & 22.4 & 43.6 & 53.8 & 16.4 & 30.4 & 67.6 & 84.7 & 19.1 & 35.8 & 70.2 & 83.1 & 15.2 & 25.1 & 49.8 & 60.8 \\ \hline
      STS-DeFee(MLP)-r1 & 18.4 & 34.2 & 71.2 & 85.4 & 13.5 & 21.9 & 42.1 & 52.5 & 16.6 & 30.4 & 66.2 & 83.2 & 19.5 & 35.5 & 69.4 & 82.3 & 15.0 & 24.8 & 49.6 & 59.8 \\
      \rowcolor[HTML]{E8E9FF} 
      STS-DeFee(MLP)-r2 & \textbf{16.4} & \textbf{32.8} & \textbf{68.8} & \textbf{82.5} & \textbf{11.4} & \textbf{19.8} & \textbf{40.8} & \textbf{49.7} & \textbf{14.8} & \textbf{28.3} & \textbf{64.8} & \textbf{80.5} & \textbf{16.6} & \textbf{32.5} & \textbf{67.6} & \textbf{80.6} & \textbf{14.0} & \textbf{23.3} & \textbf{47.5} & \textbf{58.7} \\
      STS-DeFee(GRU)-r1 & 18.5 & 34.5 & 71.4 & 85.8 & 13.5 & 22.1 & 42.5 & 53.0 & 16.2 & 30.2 & 66.7 & 83.5 & 18.9 & 35.2 & 69.4 & 82.4 & 15.1 & 24.9 & 49.5 & 59.6 \\
      \rowcolor[HTML]{E8E9FF} 
      STS-DeFee(GRU)-r2 & \textbf{16.8} & \textbf{33.0} & \textbf{68.5} & \textbf{83.2} & \textbf{11.6} & \textbf{19.9} & \textbf{41.0} & \textbf{50.1} & \textbf{14.7} & \textbf{28.3} & \textbf{65.0} & \textbf{81.1} & \textbf{16.8} & \textbf{32.7} & \textbf{67.9} & \textbf{80.3} & \textbf{14.2} & \textbf{23.6} & \textbf{47.7} & \textbf{58.7} \\ \hline \hline
       & \multicolumn{4}{c|}{sittingdown} & \multicolumn{4}{c|}{takingphoto} & \multicolumn{4}{c|}{waiting} & \multicolumn{4}{c|}{walkingdog} & \multicolumn{4}{c}{walkingtogether} \\
      frame num. & 2 & 4 & 8 & 10 & 2 & 4 & 8 & 10 & 2 & 4 & 8 & 10 & 2 & 4 & 8 & 10 & 2 & 4 & 8 & 10 \\ \hline
      MotionMixer \cite{motmix} & 12.0 & 31.4 & 64.4 & 74.5 & 9.0 & 18.9 & 41.0 & 51.6 & 10.2 & 21.1 & 45.2 & 56.4 & 20.5 & 42.8 & 75.6 & 87.8 & 10.5 & 20.6 & 38.7 & 43.5 \\ \hline
      MotMix-DeFee(MLP)-r1 & 12.0 & 30.6 & 63.7 & 72.6 & 9.1 & 18.2 & 39.7 & 50.4 & 10.4 & 20.5 & 44.1 & 55.6 & 19.5 & 42.4 & 74.4 & 86.9 & 10.2 & 19.8 & 37.8 & 42.7 \\
      \rowcolor[HTML]{E8E9FF} 
      MotMix-DeFee(MLP)-r2 & \textbf{9.8} & \textbf{29.1} & \textbf{61.8} & \textbf{70.2} & \textbf{7.8} & \textbf{16.9} & \textbf{37.1} & \textbf{47.7} & \textbf{9.3} & \textbf{19.5} & \textbf{42.0} & \textbf{53.4} & \textbf{17.3} & \textbf{40.9} & \textbf{72.8} & \textbf{84.4} & \textbf{8.3} & \textbf{19.1} & \textbf{35.9} & \textbf{41.5} \\
      MotMix-DeFee(GRU)-r1 & 11.7 & 30.8 & 63.6 & 73.7 & 8.8 & 18.5 & 40.2 & 50.8 & 10.1 & 20.8 & 44.5 & 55.4 & 19.5 & 42.1 & 74.4 & 87.1 & 10.4 & 19.8 & 38.0 & 43.1 \\
      \rowcolor[HTML]{E8E9FF} 
      MotMix-DeFee(GRU)-r2 & \textbf{10.1} & \textbf{29.4} & \textbf{62.0} & \textbf{70.8} & \textbf{7.8} & \textbf{16.9} & \textbf{38.3} & \textbf{47.9} & \textbf{9.6} & \textbf{19.8} & \textbf{42.3} & \textbf{53.6} & \textbf{17.6} & \textbf{41.1} & \textbf{72.7} & \textbf{84.9} & \textbf{8.8} & \textbf{19.0} & \textbf{36.1} & \textbf{41.8} \\ \hline
      \end{tabular}}
  \caption{Top to bottom: Prediction errors produced by the original baselines (isolated), baselines with MLP-based DeFeeNet inserted (round 1 and 2), and with GRU-based DeFeeNet inserted (round 1 and 2). Values in bold indicate lower errors and prove the deviation feedback is valid. For each baseline, we present the performance on 5 actions out of 15 in Human3.6M. We present prediction errors at frame 2, 4, 8, and 10.}
  \label{h36-round2}
\end{table*}

\subsection{Two-Round Training \& Multi-Round Testing}
\label{3.4}

The goal of our DeFeeNet is to improve the original prediction accuracy by consecutive perception and feedback on prediction deviation, as well as maintaining this priority in a sustained period of time. As the semi-overlapped structure offers us a natural way to learn prediction deviation, we accordingly present our resampling method and loss function.

\noindent\textbf{Resampling.}
Compared to existing prediction models that regard the ``observe then predict'' unit as a sample, we draw two rounds of unit as one training sample and multiple rounds of unit as one testing sample, i.e., $ \textbf{R}_{\text{train}} = [\textbf{R}_1, \textbf{R}_2] $ and $ \textbf{R}_{\text{test}} = [\textbf{R}_1, \textbf{R}_2, \cdots, \textbf{R}_{\max(r)}] $. For training phase, we just let DeFeeNet learn to produce useful information according to the previous adjacent deviation it percepts, so only two rounds of prediction (containing one overlapped region) are sufficient. While for testing phase, samples with multiple yet flexible rounds of prediction unit are necessary, as our ultimate goal is to evaluate whether DeFeeNet is able to generate stable improvement on the consecutive observation and prediction with unknown duration in advance.

\noindent\textbf{Loss Function.}
Following current works \cite{Mao-2019,Dang-2021,stsgcn,motmix,motmlp}, our loss is based on Mean Per Joint Position Error (MPJPE) \cite{dataset-h36}, but differently, we calculate the loss on both rounds of prediction. In line with notation in Section \ref{3.1}, we express the loss at round $ r $ $ (r =1,2) $ as:
\begin{equation}
    \mathcal{L}_r = \frac{1}{T} \sum_{t=1}^{T} \| \hat{\textbf{y}}_{r,t}-\textbf{y}_{r,t} \|_2,
  \label{loss_beginning}
\end{equation}
where $ \hat{\textbf{y}}_{r,t} $ denotes the predicted pose at frame $ t $ at round $ r $, and $ \textbf{y}_{r,t} $ as the corresponding ground truth. Note that, for fair comparison, Eq. \ref{loss_beginning} will have some slight changes in the specific details when our DeFeeNet is added on different baseline models and evaluated on different datasets (may refer to the supplemental material).

Then the total loss can be calculated as:
\begin{equation}
  \mathcal{L}_{\text{total}} = \mathcal{L}_1 + \mathcal{L}_2.
  \label{loss_total}
\end{equation}
In addition to supervising both rounds of prediction as accurate as possible, $ \mathcal{L}_1 $ could further ensure that the inclusion of DeFeeNet does not encumber the main pipeline too much (as the model parameters have been trained to accommodate the existence of DeFeeNet, but no deviation is involved in round 1), while $ \mathcal{L}_2 $ could keep the deviation feedback valid.

\section{Experiments}

In this section, we evaluate our DeFeeNet on the task of consecutive human motion prediction. Two datasets (Human3.6M \cite{dataset-h36} and the more recent and challenging BABEL \cite{dataset-babel}) are preprocessed to meet the task requirements, and we conduct experiments on three different motion prediction baselines with DeFeeNet inserted. We additionally provide ablation studies for further analysis.

\subsection{Datasets}

\noindent\textbf{Human3.6M} \cite{dataset-h36} is one of the most widely used datasets for human motion modeling, which includes seven actors performing actions under 15 categories, such as \emph{walking}, \emph{eating}, \emph{smoking}, and \emph{discussion}, with downsampling rate 25 fps to generate the sequence data. Each human skeletal pose is represented by 32 joints, and we follow existing works \cite{Mao-2019,Dang-2021,stsgcn,gating,motmix,motmlp} that evaluate on 22 of these joints. Currently, they aim to produce 10 or 25 frames (i.e., 400 ms short-term or 1000 ms long-term prediction) based on 10 frames of observation, within the isolated ``unit''.

For consecutive prediction task, we use our resampling method (Section \ref{3.4}) for further preprocessing. For fair comparison, we also set each prediction round with 10 frames observed and the following 10 to be predicted. Training samples are composed of two adjacent rounds, wherein the second round is 10 frames further compared to the first one. Testing samples are first arranged with 2 rounds to prove the deviation feedback between adjacent units is valid, then we arrange 10 rounds to evaluate the stability of DeFeeNet along a relatively long horizon.

\begin{table}[ht]
  \centering
  \renewcommand{\arraystretch}{1.1}
  \setlength\tabcolsep{3.2pt}
  \scalebox{0.54}{
    \begin{tabular}{c|cccc||cc|cc|cc}
      \hline
       & \multicolumn{4}{c||}{LTD-GCN \cite{Mao-2019}} & \multicolumn{2}{c|}{LTD-GCN \cite{Mao-2019}} & \multicolumn{2}{c|}{STS-GCN \cite{stsgcn}} & \multicolumn{2}{c}{MotionMixer \cite{motmix}} \\
      frame num. & 2 & 4 & 8 & 10 & \multicolumn{2}{c|}{avg} & \multicolumn{2}{c|}{avg} & \multicolumn{2}{c}{avg} \\ \hline
      isolated & 12.69 & 26.06 & 52.28 & 63.53 & \multicolumn{2}{c|}{38.64} & \multicolumn{2}{c|}{41.16} & \multicolumn{2}{c}{35.52} \\ \hline \hline
       & \multicolumn{4}{c||}{-DeFee(MLP)} & -D(MLP) & -D(GRU) & -D(MLP) & -D(GRU) & -D(MLP) & -D(GRU) \\ \hline
      r1 & 12.59 & 25.16 & 50.02 & 61.47 & 37.31 & 37.65 & 40.76 & 40.89 & 34.12 & 34.35 \\
      \rowcolor[HTML]{E8E9FF}
      r2 & \textbf{10.30} & \textbf{22.60} & \textbf{48.17} & \textbf{59.79} & \textbf{35.21} & \textbf{35.80} & \textbf{38.15} & \textbf{38.77} & \textbf{32.38} & \textbf{32.69} \\
      \rowcolor[HTML]{E8E9FF}
      r3 & \textbf{10.43} & \textbf{22.74} & \textbf{48.44} & \textbf{60.14} & \textbf{35.44} & \textbf{35.67} & \textbf{38.33} & \textbf{38.67} & \textbf{32.16} & \textbf{32.74} \\
      \rowcolor[HTML]{E8E9FF}
      r4 & \textbf{10.39} & \textbf{22.69} & \textbf{48.31} & \textbf{59.95} & \textbf{35.33} & \textbf{35.66} & \textbf{38.46} & \textbf{38.72} & \textbf{32.05} & \textbf{32.75} \\
      \rowcolor[HTML]{E8E9FF}
      r5 & \textbf{10.38} & \textbf{22.68} & \textbf{48.27} & \textbf{59.91} & \textbf{35.31} & \textbf{35.91} & \textbf{38.45} & \textbf{38.90} & \textbf{32.17} & \textbf{32.59} \\
      \rowcolor[HTML]{E8E9FF}
      r6 & \textbf{10.38} & \textbf{22.67} & \textbf{48.25} & \textbf{59.87} & \textbf{35.30} & \textbf{36.10} & \textbf{38.37} & \textbf{39.05} & \textbf{32.34} & \textbf{32.69} \\
      \rowcolor[HTML]{E8E9FF}
      r7 & \textbf{10.37} & \textbf{22.66} & \textbf{48.24} & \textbf{59.88} & \textbf{35.29} & \textbf{35.88} & \textbf{38.41} & \textbf{38.95} & \textbf{32.33} & \textbf{32.66} \\
      \rowcolor[HTML]{E8E9FF}
      r8 & \textbf{10.39} & \textbf{22.68} & \textbf{48.27} & \textbf{59.89} & \textbf{35.31} & \textbf{35.66} & \textbf{38.45} & \textbf{38.89} & \textbf{32.47} & \textbf{32.77} \\
      \rowcolor[HTML]{E8E9FF}
      r9 & \textbf{10.40} & \textbf{22.71} & \textbf{48.30} & \textbf{59.92} & \textbf{35.33} & \textbf{35.75} & \textbf{38.38} & \textbf{38.93} & \textbf{32.41} & \textbf{32.71} \\
      \rowcolor[HTML]{E8E9FF}
      r10 & \textbf{10.41} & \textbf{22.72} & \textbf{48.33} & \textbf{59.96} & \textbf{35.35} & \textbf{35.70} & \textbf{38.49} & \textbf{38.78} & \textbf{32.44} & \textbf{32.75} \\ \hline
      \end{tabular}}
  \caption{Consecutive 10 rounds of prediction errors on Human3.6M. Values in bold from round 2 to 10 indicate that deviation-aware prediction \emph{stably} yields improved performance. Detailed errors of LTD-DeFee(MLP) at frame 2, 4, 8, 10 are on the left. Average errors (avg) of these four testpoints produced by other baselines with DeFeeNet (abbreviated as \emph{D}) are on the right.}
  \label{h36-round10}
\end{table}

\noindent\textbf{BABEL} \cite{dataset-babel} is a recent proposed dataset with language labels on actions. Different from other datasets for human motion prediction wherein actions contained in each sequence are of the same category, BABEL allows for sequence with multiple kinds of actions. Currently, the most suitable preprocessing method for our task is \cite{Mao-2022} that exports two dataset files: (a) single-action sequences under 20 action labels (such as \emph{stand}, \emph{walk}, \emph{step} and \emph{stretch}); (b) sequences with two categories of actions and their in-between transitions that contain these 20 actions, both of which are of 30 fps, with only pose parameters for evaluation, while SMPL shape parameters discarded (may refer to the link provided in \cite{Mao-2022}).

As no experiments of deterministic motion prediction has been conducted on BABEL, we first cut isolated unit samples from the above sequence files to train the three baselines, with observed and predicted frames both set to 10. Then, similar to Human3.6M, we rearrange our training samples with 2 rounds while testing samples with 2 rounds for effectiveness validation and 5 rounds for stability validation. Moreover, we remove the actions with less than 200 sequence samples and leave samples with 11 action categories for evaluation.

\noindent\textbf{Evaluation Metric.}
Following \cite{Dang-2021,progressively,stsgcn,gating,motmix,motmlp}, we evaluate our network by Mean Per Joint Position Error (MPJPE) on 3D human joint coordinates, which calculates the average $ L_2 $-norm on discrepancy of different joints between prediction and the corresponding ground truth.

\subsection{Baselines}

Our DeFeeNet is agnostic to its basic prediction models, and can be added on many existing networks to realize consecutive human motion prediction. Here we introduce three baselines for our experiments.

\noindent\textbf{LTD-GCN} \cite{Mao-2019} is a GCN-based model composed of 12 GCN blocks depicting the spatial correlations of different joint dimensions with residual connections. Discrete Cosine Transformation (DCT) is used to extract temporal features.

\noindent\textbf{STS-GCN} \cite{stsgcn} designs the single-graph structure which allows for space-time cross-talk, which factorizes the graph into separable space and time matrices to fully learn the joint-joint and time-time interactions.

\noindent\textbf{MotionMixer} \cite{motmix} proposes efficient motion prediction model that only adopts MLPs by sequentially mixing both spatial and temporal dependencies, and with squeeze-and-excitation blocks enhanced.

\begin{table}[ht]
  \centering
  \renewcommand{\arraystretch}{1.1}
  \setlength\tabcolsep{3.2pt}
  \scalebox{0.60}{
    \begin{tabular}{c|cccc||cccc}
      \hline
       & \multicolumn{4}{c||}{w/o transi} & \multicolumn{4}{c}{w/ transi} \\ \hline
      frame num. & 3 & 6 & 8 & 10 & 3 & 6 & 8 & 10 \\ \hline
      LTD-GCN \cite{Mao-2019} & 0.2111 & 0.3828 & 0.4751 & 0.5617 & 0.2179 & 0.3961 & 0.4940 & 0.5802 \\ \hline
      LTD-D(MLP)-r1 & 0.2032 & 0.3781 & 0.4689 & 0.5559 & 0.2054 & 0.3805 & 0.4811 & 0.5681 \\
      \rowcolor[HTML]{E8E9FF} 
      LTD-D(MLP)-r2 & \textbf{0.1971} & \textbf{0.3683} & \textbf{0.4566} & \textbf{0.5464} & \textbf{0.1988} & \textbf{0.3699} & \textbf{0.4614} & \textbf{0.5481} \\
      LTD-D(GRU)-r1 & 0.2009 & 0.3778 & 0.4762 & 0.5601 & 0.2022 & 0.3792 & 0.4799 & 0.5614 \\
      \rowcolor[HTML]{E8E9FF} 
      LTD-D(GRU)-r2 & \textbf{0.1921} & \textbf{0.3637} & \textbf{0.4583} & \textbf{0.5470} & \textbf{0.1945} & \textbf{0.3708} & \textbf{0.4607} & \textbf{0.5434} \\ \hline \hline
      STS-GCN \cite{stsgcn} & 0.2404 & 0.4283 & 0.5330 & 0.6306 & 0.2394 & 0.4323 & 0.5536 & 0.6438 \\ \hline
      STS-D(MLP)-r1 & 0.2277 & 0.4063 & 0.5164 & 0.6063 & 0.2274 & 0.4064 & 0.5061 & 0.5960 \\
      \rowcolor[HTML]{E8E9FF} 
      STS-D(MLP)-r2 & \textbf{0.2101} & \textbf{0.3809} & \textbf{0.4933} & \textbf{0.5745} & \textbf{0.2147} & \textbf{0.3855} & \textbf{0.4839} & \textbf{0.5760} \\
      STS-D(GRU)-r1 & 0.2291 & 0.4138 & 0.5119 & 0.6140 & 0.2279 & 0.4126 & 0.5107 & 0.6028 \\
      \rowcolor[HTML]{E8E9FF} 
      STS-D(GRU)-r2 & \textbf{0.2143} & \textbf{0.3903} & \textbf{0.4917} & \textbf{0.5885} & \textbf{0.2163} & \textbf{0.3942} & \textbf{0.4927} & \textbf{0.5866} \\ \hline \hline
      MotionMixer \cite{motmix} & 0.1975 & 0.3688 & 0.4401 & 0.5172 & 0.1968 & 0.3754 & 0.4591 & 0.5386 \\ \hline
      MotMix-D(MLP)-r1 & 0.1926 & 0.3615 & 0.4372 & 0.5119 & 0.1905 & 0.3644 & 0.4426 & 0.5178 \\
      \rowcolor[HTML]{E8E9FF} 
      MotMix-D(MLP)-r2 & \textbf{0.1763} & \textbf{0.3443} & \textbf{0.4140} & \textbf{0.4961} & \textbf{0.1791} & \textbf{0.3511} & \textbf{0.4268} & \textbf{0.4989} \\
      MotMix-D(GRU)-r1 & 0.1884 & 0.3617 & 0.4407 & 0.5136 & 0.1899 & 0.3685 & 0.4422 & 0.5151 \\
      \rowcolor[HTML]{E8E9FF} 
      MotMix-D(GRU)-r2 & \textbf{0.1743} & \textbf{0.3512} & \textbf{0.4223} & \textbf{0.4881} & \textbf{0.1768} & \textbf{0.3541} & \textbf{0.4248} & \textbf{0.4896} \\ \hline
      \end{tabular}}
  \caption{Comparisons of prediction errors between the original baselines (isolated) and baselines with MLP/GRU-based DeFeeNet (round 1 and round 2). We conduct experiments on BABEL with (a) only single-action samples (i.e., w/o transi) and (b) two-action samples with transitions between different actions (i.e., w/ transi). Lower errors in round 2 (marked in bold) show that the deviation produced previously is effective to improve the subsequent prediction accuracy. We abbreviate our DeFeeNet as \emph{D}.}
  \label{babel-round2}
\end{table}

\subsection{Implementation Details}

We conduct our experiments under Pytorch \cite{pytorch} framework with Adam optimizer \cite{adam} on a single NVIDIA RTX 2080Ti. In MLP-based version, the hidden layer size of temporal-mixing and spatial-mixing is set to the same as the corresponding dimensions of the basic pipeline. In GRU-based version, the hidden size is set to 256 regardless of the basic model, while fully-connected layers are used for dimension alignment. Details are provided in the supplemental material. For LTD-GCN, STS-GCN, and MotionMixer, we jointly train each baseline with our DeFeeNet with learning rate as $ 0.0005 $, $ 0.01 $ and $ 0.01 $, respectively. We train 50 epochs on Human3.6M and 100 on BABEL.

\begin{table*}[ht]
  \centering
  \renewcommand{\arraystretch}{1.1}
  \setlength\tabcolsep{3.2pt}
  \scalebox{0.64}{
    \begin{tabular}{c|cccccc||cccccccccc}
      \hline
      \multicolumn{1}{l|}{} & \multicolumn{6}{c||}{w/o transi} & \multicolumn{10}{c}{w/ transi} \\ \hline
      \multicolumn{1}{l|}{} & \multicolumn{2}{c|}{LTD-GCN \cite{Mao-2019}} & \multicolumn{2}{c|}{STS-GCN \cite{stsgcn}} & \multicolumn{2}{c||}{MotionMixer \cite{motmix}} & \multicolumn{2}{c|}{LTD-GCN \cite{Mao-2019}} & \multicolumn{2}{c|}{STS-GCN \cite{stsgcn}} & \multicolumn{2}{c||}{MotionMixer \cite{motmix}} & \multicolumn{4}{c}{STS-GCN \cite{stsgcn}} \\
      \multicolumn{1}{l|}{frame num.} & \multicolumn{2}{c|}{avg} & \multicolumn{2}{c|}{avg} & \multicolumn{2}{c||}{avg} & \multicolumn{2}{c|}{avg} & \multicolumn{2}{c|}{avg} & \multicolumn{2}{c||}{avg} & 3 & 6 & 8 & 10 \\ \hline
      isolated & \multicolumn{2}{c|}{0.4077} & \multicolumn{2}{c|}{0.4581} & \multicolumn{2}{c||}{0.3809} & \multicolumn{2}{c|}{0.4221} & \multicolumn{2}{c|}{0.4673} & \multicolumn{2}{c||}{0.3925} & 0.2394 & 0.4323 & 0.5536 & 0.6438 \\ \hline \hline
      \multicolumn{1}{l|}{} & -D(MLP) & \multicolumn{1}{c|}{-D(GRU)} & -D(MLP) & \multicolumn{1}{c|}{-D(GRU)} & -D(MLP) & -D(GRU) & -D(MLP) & \multicolumn{1}{c|}{-D(GRU)} & -D(MLP) & \multicolumn{1}{c|}{-D(GRU)} & -D(MLP) & \multicolumn{1}{c||}{-D(GRU)} & \multicolumn{4}{c}{-DeFee(MLP)} \\ \hline
      r1 & 0.4015 & \multicolumn{1}{c|}{0.4038} & 0.4392 & \multicolumn{1}{c|}{0.4421} & 0.3758 & 0.3761 & 0.4089 & \multicolumn{1}{c|}{0.4055} & 0.4413 & \multicolumn{1}{c|}{0.4385} & 0.3785 & \multicolumn{1}{c||}{0.3766} & 0.2372 & 0.4112 & 0.5139 & 0.6029 \\
      \rowcolor[HTML]{E8E9FF} 
      r2 & \textbf{0.3842} & \multicolumn{1}{c|}{\cellcolor[HTML]{E8E9FF}\textbf{0.3901}} & \textbf{0.4185} & \multicolumn{1}{c|}{\cellcolor[HTML]{E8E9FF}\textbf{0.4239}} & \textbf{0.3590} & \textbf{0.3589} & \textbf{0.3982} & \multicolumn{1}{c|}{\cellcolor[HTML]{E8E9FF}\textbf{0.3963}} & \textbf{0.4270} & \multicolumn{1}{c|}{\cellcolor[HTML]{E8E9FF}\textbf{0.4265}} & \textbf{0.3547} & \multicolumn{1}{c||}{\cellcolor[HTML]{E8E9FF}\textbf{0.3599}} & \textbf{0.2342} & \textbf{0.3965} & \textbf{0.4949} & \textbf{0.5824} \\
      \rowcolor[HTML]{E8E9FF} 
      r3 & \textbf{0.3810} & \multicolumn{1}{c|}{\cellcolor[HTML]{E8E9FF}\textbf{0.3890}} & \textbf{0.4122} & \multicolumn{1}{c|}{\cellcolor[HTML]{E8E9FF}\textbf{0.4203}} & \textbf{0.3611} & \textbf{0.3566} & \textbf{0.3815} & \multicolumn{1}{c|}{\cellcolor[HTML]{E8E9FF}\textbf{0.3802}} & \textbf{0.4205} & \multicolumn{1}{c|}{\cellcolor[HTML]{E8E9FF}\textbf{0.4189}} & \textbf{0.3502} & \multicolumn{1}{c||}{\cellcolor[HTML]{E8E9FF}\textbf{0.3582}} & \textbf{0.2332} & \textbf{0.3920} & \textbf{0.4845} & \textbf{0.5723} \\
      \rowcolor[HTML]{E8E9FF} 
      r4 & \textbf{0.3808} & \multicolumn{1}{c|}{\cellcolor[HTML]{E8E9FF}\textbf{0.3826}} & \textbf{0.4132} & \multicolumn{1}{c|}{\cellcolor[HTML]{E8E9FF}\textbf{0.4277}} & \textbf{0.3582} & \textbf{0.3540} & \textbf{0.3877} & \multicolumn{1}{c|}{\cellcolor[HTML]{E8E9FF}\textbf{0.3845}} & \textbf{0.4223} & \multicolumn{1}{c|}{\cellcolor[HTML]{E8E9FF}\textbf{0.4192}} & \textbf{0.3610} & \multicolumn{1}{c||}{\cellcolor[HTML]{E8E9FF}\textbf{0.3570}} & \textbf{0.2316} & \textbf{0.3904} & \textbf{0.4897} & \textbf{0.5777} \\
      \rowcolor[HTML]{E8E9FF} 
      r5 & \textbf{0.3855} & \multicolumn{1}{c|}{\cellcolor[HTML]{E8E9FF}\textbf{0.3865}} & \textbf{0.4166} & \multicolumn{1}{c|}{\cellcolor[HTML]{E8E9FF}\textbf{0.4359}} & \textbf{0.3545} & \textbf{0.3578} & \textbf{0.3840} & \multicolumn{1}{c|}{\cellcolor[HTML]{E8E9FF}\textbf{0.3866}} & \textbf{0.4290} & \multicolumn{1}{c|}{\cellcolor[HTML]{E8E9FF}\textbf{0.4133}} & \textbf{0.3579} & \multicolumn{1}{c||}{\cellcolor[HTML]{E8E9FF}\textbf{0.3613}} & \textbf{0.2330} & \textbf{0.3933} & \textbf{0.4952} & \textbf{0.5946} \\ \hline
      \end{tabular}}
  \caption{The consecutive 5 rounds of prediction errors on BABEL (w/o or w/ transi). The detailed errors of STS-DeFee(MLP) at frame 3, 6, 8, 10 are presented on the right. Average errors (avg) of these four testpoints produced by other baselines with DeFeeNet (abbreviated as \emph{D}) are also provided. From round 2 to 5, our method could \emph{stably} yield lower prediction errors than isolated baselines (marked in bold).}
  \label{babel-round5}
\end{table*}

\begin{figure*}[t]
  \centering
   \includegraphics[width=0.94\linewidth]{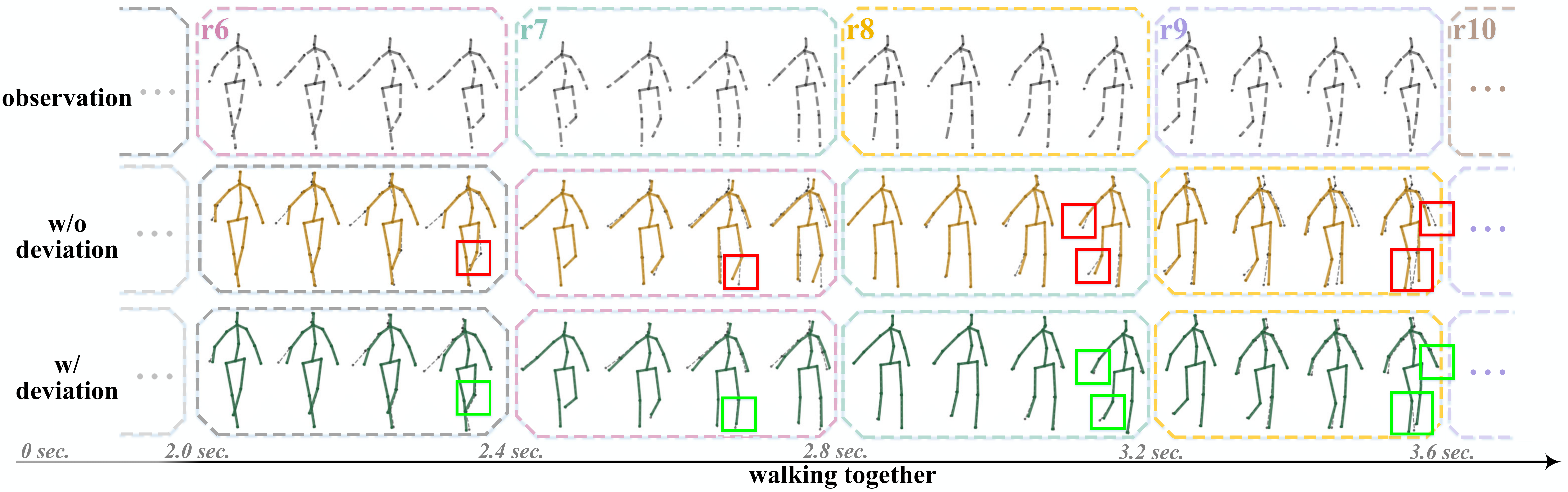}

   \caption{Visualized comparisons of consecutive motion prediction w/o or w/ deviation feedback on a sample \emph{walking together} from Human3.6M. Due to the limited space, we only provide 3 prediction rounds (r6, r7, r8) out of 10, with each round marked by a unique color (details are provided in Section \ref{4.4}). Here the original predictor is LTD-GCN \cite{Mao-2019}, and we add our MLP-based DeFeeNet on it. From top to bottom: observation/GT (covering a relatively long horizon of motions for consecutive prediction), w/o deviation (directly employing the existing predictor round by round to roughly implement consecutive prediction), w/ deviation (deviation-aware consecutive prediction with DeFeeNet). 
   Poses drawn in dashed lines indicate GT, and our DeFeeNet helps to yield improved prediction.}
   \label{fig:walkingtogether}
\end{figure*}

\begin{figure*}[t]
  \centering
   \includegraphics[width=0.94\linewidth]{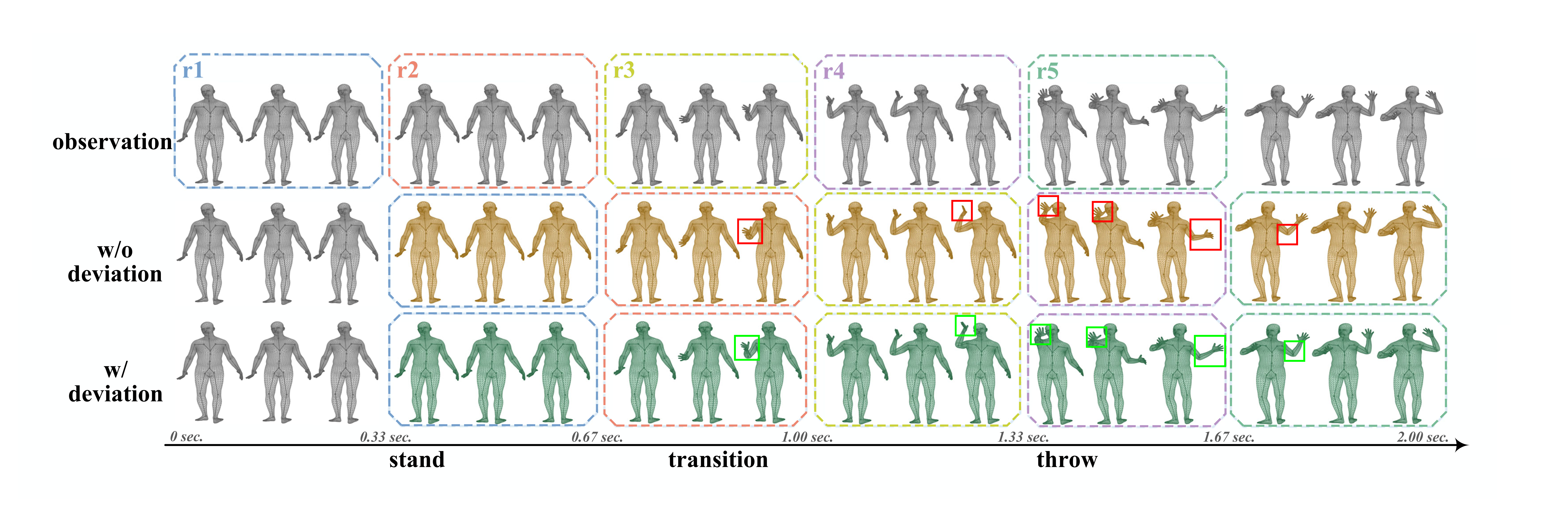}

   \caption{Visualized comparisons of consecutive motion prediction w/o or w/ deviation feedback on a sample \emph{stand-transition-throw} from BABEL. Each round is marked with a unique color. Here the original predictor is STS-GCN \cite{stsgcn} to roughly implement consecutive prediction. We add our GRU-based DeFeeNet on it, and obtain the improved prediction results with more accurate and elegant motions.}
   \label{fig:walk-transi-throw}
\end{figure*}

\subsection{Results on Human3.6M}
\label{4.4}

Table \ref{h36-round2} presents the prediction errors of models with and without DeFeeNet inserted on Human3.6M dataset. Without DeFeeNet, current baselines observe and predict human motions within the isolated short unit (i.e., length of 10 frames is 400 ms). However, we jointly train two rounds of prediction process, wherein the prediction deviation generated in round 1 (r1) is detected and learned by our inserted DeFeeNet, in order to assist the original pipeline to predict with this previous deviation as feedback to improve the performance of round 2 (highlighted in purple background color). Values in bold indicate lower errors, which prove that this deviation information between adjacent prediction rounds is valid for prediction improvement.

To demonstrate the stability of DeFeeNet for consecutive human motion prediction, we further show the results of consecutive 10 rounds of prediction in Table \ref{h36-round10}. From errors of round 2 to 10, we observe that, compared to baselines which treat motion prediction as a task within the isolated unit, our deviation-aware prediction manner could \emph{stably} produce improved results during a sustained period of time. Figure \ref{fig:walkingtogether} shows visualized comparisons of consecutive motion prediction w/o or w/ deviation feedback on a sample of action \emph{walking together}. We provide r6, r7 and r8 out of the total 10 rounds that marked with different colors. For example, in r8, the existing predictor only takes as input the poses in the yellow box from \emph{line GT} as observation, and then predict the poses in the yellow box from \emph{line w/o deviation}, to roughly implement one round of prediction. However, our DeFeeNet is able to detect the deviation between current observation and previous prediction of r7 (poses in the mint green from \emph{line w/ deviation}), and therefore use this deviation feedback to predict more accurately (see poses in the yellow box from \emph{line w/deviation}). Meanwhile, these predicted poses could also help the next round of prediction (r9) in the same manner as above. Note that we only draw 4 poses in each observed/predicted box for space saving, and the actual pose number is 10.

Additionally, in both Table \ref{h36-round2} and \ref{h36-round10}, we find that round 1 is also improved although there is no deviation involved. This is reasonable as the total loss for each sample is calculated by adding the prediction loss of round 1 and 2, which means that the second loss would also have an effect on the first prediction. 
To remove such effect,
ablation study is given in supplemental material, containing comparison between deviation feedback enabled/disabled, to further show the effectiveness of DeFeeNet on accuracy improvement.

\subsection{Results on BABEL}

We provide the prediction errors of models with and without DeFeeNet inserted on BABEL in Table \ref{babel-round2}. We conduct experiments on (a) BABEL w/o transi (only single-motion samples of 11 action categories) and (b) BABEL w/ transi (involving two-action samples and the in-between action transitions within the 11 categories). From the table, our DeFeeNet is effective for error reduction whether action changing is involved or not. Similar to Human3.6M, the deviation produced by round 1 is learned by DeFeeNet and therefore can be used to improve the next round of prediction, which lays the foundation for consecutive motion prediction that requires deviation feedbacks round by round.

We also present the stability validation results in Table \ref{babel-round5}, which shows the errors of consecutive 5-round prediction. Although slightly fluctuating, our DeFeeNet is still able to constantly get lower errors than isolated baselines. We provide the visualized comparisons on a sample named \emph{stand-transition-throw} in Figure \ref{fig:walk-transi-throw}. As shown in the figure, our deviation-aware consecutive prediction could yield accurate prediction results even when faced with challenging action-changing samples. This is reasonable, as DeFeeNet is more sensitive to local information, i.e., the prediction deviation that \emph{just happened}, which is a potential factor of action state changes. In our experiments, 5 rounds of prediction with observation part included accounts for 60 frames (i.e., 2 seconds). We do not select a longer horizon (like Human3.6M) due to that many motion sequences in \cite{Mao-2022} are shorter than 50 frames and cannot be processed as test samples for our multi-round prediction.

\subsection{Ablation Study}

To further validate our pipeline-inserted design is effective yet tailored to human motion prediction, and prove its essential difference against the ``residual correction'' spirit (mentioned in Section \ref{2.2}), we modify our DeFeeNet as DeFee-corr: a post-processing step of existing human motion predictors that aims to correct the already-produced prediction. The main architecture is retained but moved to the end of existing predictors, which detects \emph{previous} prediction deviation, and then maps it into an estimated deviation embedding to be directly added on \emph{current} round of prediction as ``correction''. From Table \ref{defee-corr}, DeFee-corr fails to yield the improvement like ours, as baselines inserted with DeFeeNet allows for joint decoding on both current observation and previous deviation information, which leverages the decoding power of the existing predictors and therefore predict naturally improved motions, while simply adding the deviation vector produced by the simple yet small DeFee-corr for prediction correction is not suitable for delicate human motion data.

\begin{table}[ht]
  \centering
  \renewcommand{\arraystretch}{1.1}
  \setlength\tabcolsep{3.2pt}
  \scalebox{0.80}{
    \begin{tabular}{c|cc|cc|cc}
      \hline
       & \multicolumn{2}{c|}{LTD-GCN \cite{Mao-2019}} & \multicolumn{2}{c|}{STS-GCN \cite{stsgcn}} & \multicolumn{2}{c}{MotionMixer \cite{motmix}} \\ \hline
      isolated & \multicolumn{2}{c|}{0.4221} & \multicolumn{2}{c|}{0.4673} & \multicolumn{2}{c}{0.3925} \\ \hline \hline
       & \multicolumn{1}{c|}{MLP} & GRU & \multicolumn{1}{c|}{MLP} & GRU & \multicolumn{1}{c|}{MLP} & GRU \\ \hline
      DeFee-r2 & \multicolumn{1}{c|}{\textbf{0.3945}} & \textbf{0.3923} & \multicolumn{1}{c|}{\textbf{0.4150}} & \textbf{0.4224} & \multicolumn{1}{c|}{\textbf{0.3640}} & \textbf{0.3613} \\ \hline
      DeFee-corr-r2 & \multicolumn{1}{c|}{0.4187} & 0.4122 & \multicolumn{1}{c|}{0.4472} & 0.4603 & \multicolumn{1}{c|}{0.3838} & 0.3892 \\ \hline
      \end{tabular}}
  \caption{Comparisons of round 2 prediction errors (average on frame 3, 6, 8, 10) between DeFeeNet and DeFee-corr on BABEL w/ transi. Values in bold indicate lower errors.}
  \label{defee-corr}
\end{table}



\section{Conclusion}

We reformulate current human motion prediction task from the consecutive perspective which covers multiple rounds of ``observe then predict'', and argue that each round of prediction accuracy can be improved if the prediction \emph{deviation} generated in previous round is well detected and learned by robots/machines for practical use. In this paper, we propose DeFeeNet with MLP-based version and GRU-based version, both of which are simple yet effective, and can be inserted to existing human motion prediction models to realize deviation perception and feedback across adjacent prediction units. Our DeFeeNet is able to produce stably improved prediction during a relatively long horizon on Human3.6M as well as the more recent and challenging BABEL that contains samples with different actions and the in-between transitions.

\noindent\textbf{Limitations.} In real world, observations obtained by devices may contain noises and occlusions, which harm the deviation perception process, whose failure would further harm prediction quality. To utilize prediction deviation under imperfect condition still requires follow-up research.

~\\
\noindent\textbf{Acknowledgements.} This work was supported in part by the National Natural Science Foundation of China (NO. 62176125, 61772272).

{\small
\bibliographystyle{ieee_fullname}
\bibliography{egbib}
}

\end{document}